\documentclass[11pt]{article}

\usepackage[preprint]{acl}

\usepackage{times}
\usepackage{latexsym}

\usepackage[T1]{fontenc}

\usepackage[utf8]{inputenc}

\usepackage{microtype}

\usepackage{inconsolata}

\usepackage{graphicx}

\usepackage{hyperref}       
\usepackage{url}            
\usepackage{booktabs}       
\usepackage{amsfonts}       
\usepackage{nicefrac}       
\usepackage{xcolor}         
\usepackage{colortbl}       
\usepackage{multirow}       

\title{
     UniMoCo: Unified Modality Completion for Robust Multi-Modal Embeddings
}

\author{%
  Jiajun Qin$^{1*}$\quad
  Yuan Pu$^{1,2}$\thanks{Equal contribution.}\quad
  Zhuolun He$^{1,2}$\quad
  Seunggeun Kim$^3$\quad
  David Z. Pan$^3$\quad
  Bei Yu$^{1}$\\
  $^1$The Chinese University of Hong Kong, China \quad
  $^2$ChatEDA Tech \quad \\
  $^3$University of Texas at Austin, USA \\
  \texttt{hobbitqia@gmail.com}, \texttt{\{1155124579, zlhe, byu\}@cse.cuhk.edu.hk} \\
  \texttt{\{sgkim, dpan\}@utexas.edu}
}

\newcommand{\model}{UniMoCo}

\begin{document}
\maketitle

\begin{abstract}
    Current vision-language models have been explored for multi-modal embedding tasks like information retrieval. However, they face significant challenges in real-world queries and targets involving diverse modality combinations, as existing approaches often fail to align all modality combinations within a unified embedding space during training, leading to degraded performance on rare modality patterns during inference. To address this fundamental limitation, we propose UniMoCo, a novel architecture featuring a modality-completion module that generates visual features from text, thereby ensuring modality completeness for both queries and targets. Additionally, UniMoCo incorporates a specialized training strategy that aligns embeddings from both original and modality-completed inputs, thus ensuring consistent and robust embeddings for diverse modality combinations. Comprehensive experiments demonstrate that UniMoCo outperforms previous methods while exhibiting consistent robustness across diverse settings. Furthermore, we identify and quantify the inherent bias in conventional approaches caused by imbalanced modality combinations in training data, showing that our modality-completion paradigm effectively mitigates this limitation.
\end{abstract}
\section{Introduction}
\label{sec:intro}


Multi-modal embedding methods encode inputs with different modalities (such as text and image) into representations in an unified high-dimensional vector space, facilitating downstream tasks such as image classification~\cite{deng2009imagenet}, information retrieval~\cite{datta2008image,gordo2016deep}, retrieval augmented generation~\cite{zhao2023retrieving}, visual-language alignment~\cite{plummer2015flickr30k,lin2014microsoft}, etc.
Previous models such as CLIP~\cite{radford2021learning}, BLIP~\cite{li2022blip}, SigLIP~\cite{zhai2023sigmoid} and ALIGN~\cite{jia2021scaling} aim to learn unified multi-modal representations by aligning visual and textual modalities through large-scale pretraining on paired image-text data, enabling cross-modal understanding and multi-modal embedding task applications.
However, these models usually adopt the dual-encoder architecture with shallow or even no fusion of the visual and textual features,
making fine-grained cross-modal reasoning (e.g., spatial relationships or detailed text-image interactions) less effective,
limiting their application in complicated multi-modal embedding scenarios.

Recently with the rapid advancement of large vision language models (LVLMs)
\cite{zhu2023minigpt,liu2023visual,chen2024internvl,wang2024qwen2},
the extraordinary visual-textual understanding and reasoning capabilities of LVLMs have been unleashed for multi-modal representation learning and embedding tasks adaption.
VLM2VEC~\cite{jiang2024vlm2vec} introduced massive multimodal embedding benchmark (MMEB),
a comprehensive benchmark for multi-modal embedding tasks covering classification, retrieval, vision question answering (VQA) and visual grounding.
Other works propose specific training strategies~\cite{liu2024lamra,gu2025breaking,lan2025llave,zhou2024vista} or data augmentation techniques~\cite{lin2024mm,zhou2024megapairs} to train LVLMs for embedding adaption.
Despite these advancements, adapting LVLMs for embedding tasks reveals limited performance in real-world applications, where queries and targets often feature diverse and incomplete modality combinations. The challenge of missing text is effectively addressed through the use of task-specific instructions, a method supported by prior work~\cite{jiang2024vlm2vec}, and by the near-ubiquity of textual system prompts in practice. Therefore, the absence of the visual modality emerges as the more frequent and critical challenge.

Consequently, current methods struggle to align all modality combinations into a semantically coherent and unifiedembedding space during training. This misalignment, largely due to imbalanced combinations in training data, degrades performance when the model encounters underrepresented combinations during inference.

\begin{figure*}
    \centering
    \includegraphics[width=1\linewidth]{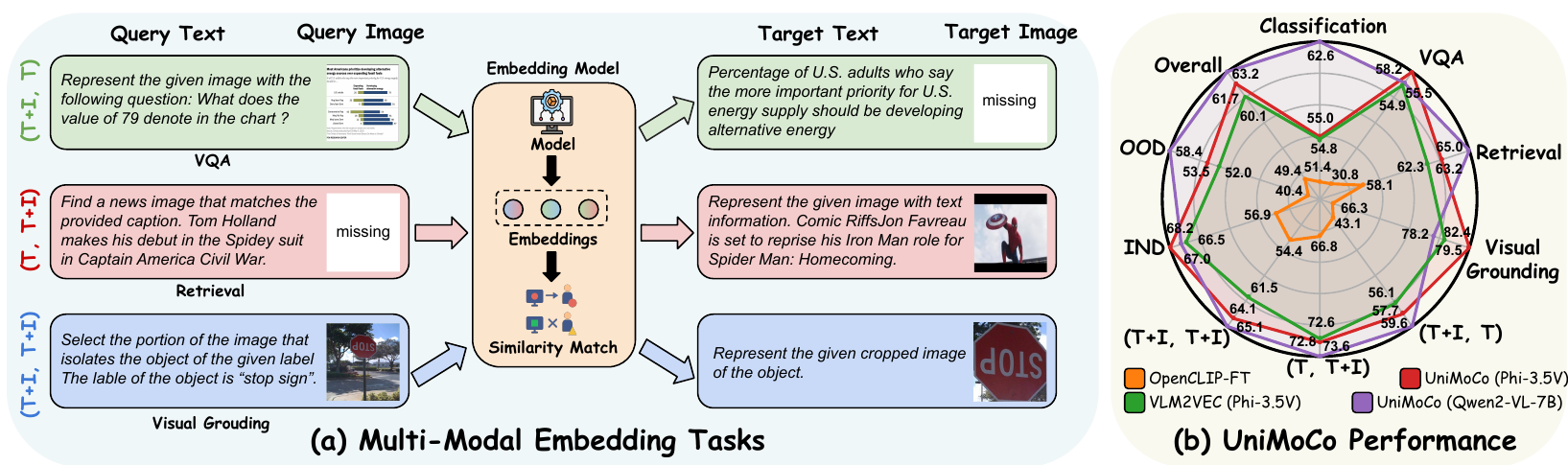}
   \caption{
    (a) Multi-modal embedding tasks involve three common modality combinations, illustrated with examples from the MMEB~\cite{jiang2024vlm2vec}: $(\mathrm{T}+\mathrm{I},\mathrm{T})$, $(\mathrm{T},\mathrm{T}+\mathrm{I})$, and $(\mathrm{T}+\mathrm{I},\mathrm{T}+\mathrm{I})$.
    Specifically, $(\mathrm{T}+\mathrm{I},\mathrm{T})$ denotes a query containing both text and image, while the target is text-only; the other combinations follow similarly.
    A multi-modal embedding model encodes both query and target into a unified space and then conduct tasks through similarity matching.
    (b) UniMoCo's performance vs. other embedding models on MMEB.
   }
    \label{fig:scenario}
\end{figure*}

In this work, we propose \textbf{UniMoCo}, a model architecture with \underline{Uni}fied \underline{Mo}dality \underline{Co}mpletion designed to learn robust multi-modal embeddings. It consists of two main components: a backbone LVLM and a modality-completion module. During training and inference, the modality-completion module synthesizes visual embeddings directly from the input text whenever the visual modality is absent, which ensures that multi-modal representations remain complete even under modality absence. Moreover, we design a complementary training strategy that integrates contrastive loss to enhance representation quality by bringing matching query-target pairs closer while pushing non-matching ones apart, and employs auxiliary loss to maintain consistency between pseudo and real visual embeddings, thereby improving the reliability of synthesized embeddings when visual inputs are missing. Together, the UniMoCo architecture and its tailored training approach enable effective handling of diverse modality combinations in the embedding space, boosting the model's adaptability and robustness in real multi-modal scenarios. Our contributions can be summarized as follows:
\begin{itemize}
    \item We propose UniMoCo, a LVLM-based framework augmented with a modality-completion module to produce robust multi-modal embeddings for diverse embedding tasks.
    \item We develop an effective training strategy combining contrastive learning with auxiliary losses to maximize UniMoCo's potential. 
    \item We evaluate our method on MMEB, which shows existing methods are prone to modality combination bias while UniMoCo not only mitigates this issue but also delivers superior performance across multiple tasks.
\end{itemize}
\section{Related Work}
\label{sec:related_work}

\subsection{Text Representation Learning}
Text embeddings are extensively utilized across diverse natural language processing tasks, including text classification, retrieval, and QA.
Current approaches to learning text representation can be broadly categorized into task-specific and general-purpose paradigms.
Early research primarily focused on developing specialized architectures for distinct applications:
works~\cite{huang2019knowledge, karpukhin2020dense, hao2019exploiting} targeted QA systems,
while \cite{wang2018joint, sinoara2019knowledge} addressed classification tasks,
and \cite{huang2020embedding, cakaloglu2020text, zhan2020repbert} specialized in retrieval scenarios.
Recent advancements have shifted toward developing general-purpose embedding models with broader applicability.
Multiple studies~\cite{giorgi2020declutr, wang2022text, li2023towards, behnamghader2024llm2vec} have successfully employed contrastive learning frameworks for this objective.
Concurrently, innovative approaches~\cite{su2022one, asai2022task} incorporate task-specific instructions alongside input text during encoding, enabling unified handling of multiple downstream tasks.
Recent research has extended the application scope of decoder-only large language models (LLMs) traditionally used for generation,
with several studies successfully utilized them as embedding models \cite{wang2023improving, li2023towards, behnamghader2024llm2vec, lee2024nv} with promising results.

\subsection{Multi-modal Representation Learning}
Unlike text embedding, multi-modal representations enable broader applicability across diverse tasks~\cite{kim2020mule, khare2021mmbert, ge2021structured, lin2022multi, xia2023graph, li2024groundinggpt, zhou2024vista}, yet their learning poses greater challenges due to the complexity of aligning different modalities. Prior approaches predominantly leverage encoder-based architectures such as CLIP~\cite{radford2021learning} and BLIP~\cite{li2022blip} to project different modalities into a unified space to align multi-modal inputs. Recent advances like VLMVec~\cite{jiang2024vlm2vec} introduce LVLMs as backbones for a more generalized framework. Subsequent advancements focus on refining contrastive learning objectives~\cite{lan2025llave, yu2025cafe}, enhancing data quality via synthetic datasets~\cite{zhou2024megapairs, chen2025mme5}, or optimizing training strategies~\cite{liu2024lamra, gu2025breaking,zhou2024vista}. However, these efforts primarily target training methodologies rather than architectural innovations and overlook the critical limitation of incomplete modality combinations, which our work systematically addresses through novel structural improvements.

\subsection{Modality Missing}
Real-world multi-modal applications often suffer from missing modalities during training or inference, which can significantly degrade performance~\cite{wu2024missingmodality_survey}.
Early dual-encoder models like CLIP, BLIP, and FLAVA~\cite{radford2021learning,li2022blip,2022flava} tackle this by learning a shared modality-invariant space.
However, their reliance on complete pre-training data introduces biases toward dominant modality combinations~\cite{ma2022missingmodality, ma2021SMIL}.

Generative approaches synthesize proxy modalities to reconstruct missing inputs~\cite{zhou2024megapairs, chen2025mme5, feng2024missingmodality_t2i},
yet the quality of the approaches is set by the off‑the‑shelf frozen generators.
Specialized expert-based methods like Flex-MoE~\cite{yun2024flexmoe} use dynamic routing,
while transformer-based approaches~\cite{jiang2024vlm2vec, lee2023promptadapter, yang2024mma} leverage prompt or adapter tuning for flexible inputs.
However, fixed dropout strategies of these methods often fail to generalize to rare modality combinations.
In contrast, we propose a lightweight modality-completion module integrated with a unified LVLM backbone,
synthesizing missing visual embeddings to ensure consistent alignment across all modality combinations.

\section{Methodology}
\label{sec:method}

\begin{figure*}[tb!]
    \centering
    \includegraphics[width=1\linewidth]{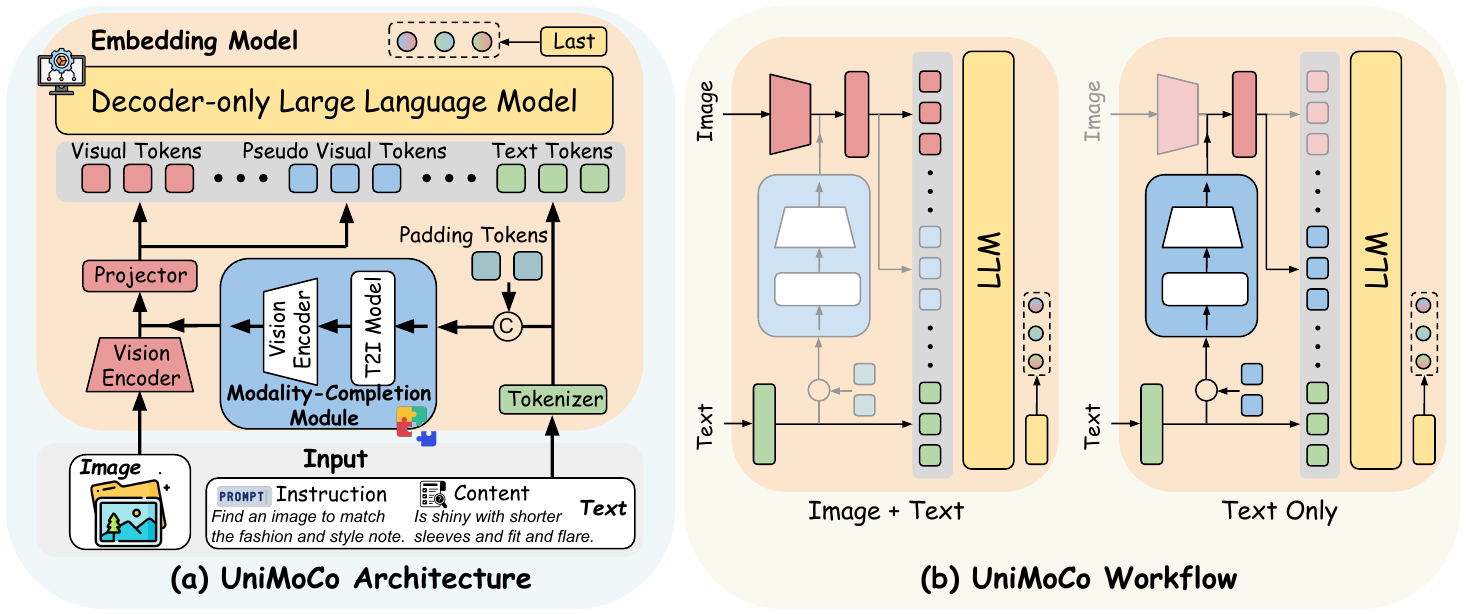}
    \caption{(a) UniMoco architecture. Processes image/text inputs through an LLM, with the final output token as the unified embedding. (b) UniMoCo workflow illustration. The left panel shows image-text processing while the right panel shows text-only input processing. Grayed-out icons indicate inactive modules in each scenario. This unified workflow supports both training and inference phases.}
    \label{fig:unimoco}
\end{figure*}


\subsection{Problem Definition}


In this work, we propose UniMoCo, a unified multi-modal embedding model that projects both the query and candidate targets into a shared high-dimensional embedding space.
Our model accepts both textual and optional visual inputs, encoding them into compact and expressive embeddings $E \in \mathbb{R}^d$,
where $d$ represents the embedding dimension.
These unified representations are designed to capture discriminative features. Within this space, similarity matching is performed to identify the candidate target embedding closest to the query embedding, making it the most suitable match.

The input data (for both query and target) comprises two elements:
(1) text, which includes a task-defining \textbf{instruction} (e.g., ``\textit{Find an image matching the fashion image and style note.}'')
and specific \textbf{content} (e.g., ``\textit{Shiny silver material with short sleeves and a fit-and-flare silhouette.}'');
(2) optional images. They are processed simultaneously to generate embeddings.

Given a query $q$ and $n$ candidate targets $\{c_1, c_2, \dots, c_n\}$, we compute their respective embeddings $E_q = f(q)$ and $E_{c_i} = f(c_i)$ for $i = 1, \dots, n$, where $f(\cdot)$ represents the embedding model. The optimal match $c^*$ is determined by selecting the candidate with the highest similarity:
$$
c^* = \arg\max_{c_i} \, \mathrm{sim}(E_q, E_{c_i}),
$$
where $\mathrm{sim}(E_q, E_{c_i})$ is typically implemented as temperature-scaled cosine similarity function.

\subsection{UniMoCo Architecture}\label{sec:arch}

Figure~\ref{fig:unimoco} presents our UniMoCo architecture, which utilizes an LVLM as its backbone with three components: LLM, vision encoder, and projector.
To ensure robust handling of diverse input modality combinations, we incorporate task-specific instructions into every input. When the  textual content is missing, the instruction itself serves as the available linguistic input, thereby compensating for the absence. For cases lacking visual input, we introduce a dedicated text-to-image (T2I) module that utilizes a compact language model to directly convert text into pseudo visual embeddings. This approach avoids the computational overhead and embedding-task misalignment associated with conventional diffusion-based T2I methods~\cite{zhou2024megapairs, chen2025mme5, feng2024missingmodality_t2i}, while effectively maintaining cross-modal alignment and functional coherence across all input conditions without redundant processing.


The modality-completion module is further enhanced through the addition of a supplementary vision encoder. This architectural decision stems from our observation that embeddings produced by the T2I model exhibit incomplete consistency with those generated by the original vision encoder processing real images. The additional encoder serves to better capture and represent the characteristics of our pseudo visual embeddings. Furthermore, when processing text tokens within this module, we concatenate them with padding tokens to maintain a fixed input length that matches the number of visual tokens produced by the primary vision encoder from authentic images.\footnote{For details of the padding token design, see Appendix~\ref{sec:pad}.} A comprehensive analysis of this module's components and their contributions is provided in Section~\ref{sec:ablation}. 

Figure~\ref{fig:unimoco} details the operational workflow for various input scenarios.
When presented with complete multi-modal inputs (Case 1), the system bypasses the completion module entirely, functioning as a conventional LVLM.
In situations where image data is unavailable (Case 2), the textual input is simultaneously processed by both the LLM and our completion module,
with the latter generating pseudo visual tokens to substitute for the missing image representations.
This dual-path approach ensures robust performance across different inputs while maintaining model consistency.

\begin{figure*}[tb!]
    \centering
    \includegraphics[width=1\linewidth]{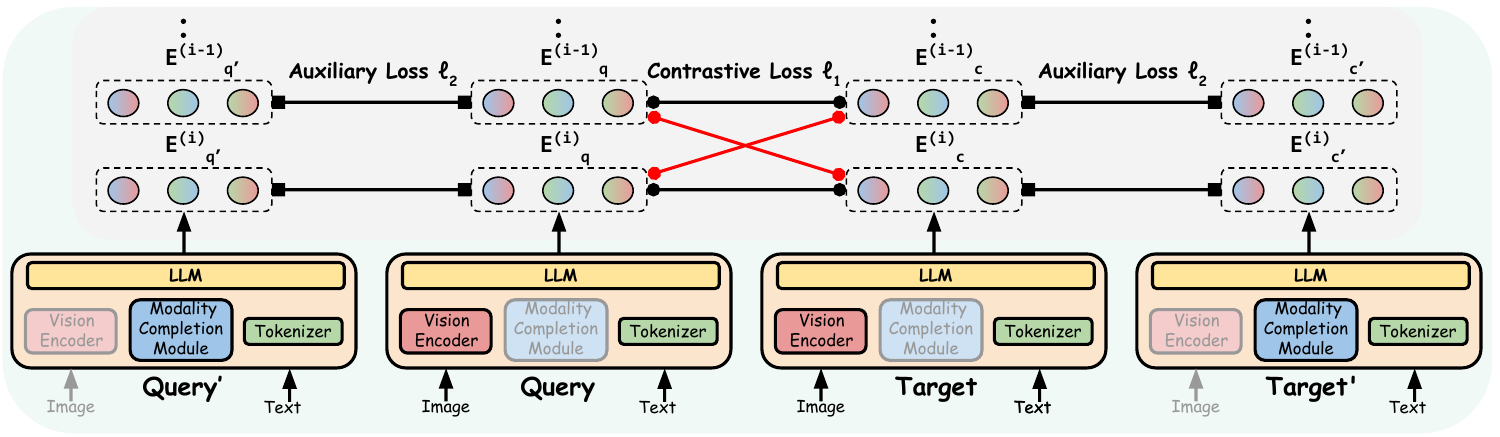}
    \caption{
        UniMoCo training combines a primary contrastive loss ($\mathcal{L}_1$) with auxiliary loss ($\mathcal{L}_2$). Black lines indicate positive pairs to be pulled closer in the embedding space, while red lines denote negative pairs requiring separation.
    }
    \label{fig:loss}
\end{figure*}

\subsection{UniMoCo Training Strategy}\label{sec:loss}

As illustrated in Figure~\ref{fig:loss}, our framework employs two complementary loss functions.
In conventional multi-modal embedding frameworks~\cite{jiang2024vlm2vec},
contrastive learning aims to learn discriminative representations by minimizing distance between query-target positive pairs while maximizing separation from negatives.
For a batch of $B$ query-target pairs, the InfoNCE loss is defined as:
\begin{equation}
\mathcal{L}_1 = -\frac{1}{B} \sum_{i=1}^B\log\left(\frac{\exp(\mathrm{sim}(E_q^{(i)}, E_{c^+}^{(i)})/\tau}{\sum_{j=1}^B \exp(\mathrm{sim}(E_q^{(i)}, E_{c_j}^{(i)})/\tau)}\right),
\end{equation}
where $E_q^{(i)}, E_{c^+}^{(i)} \in \mathbb{R}^d$ denote the $i$-th query and positive target embeddings respectively, $B$ represents batch size, and $\tau > 0$ is the temperature. 

However, modality-complete and modality-missing inputs produce fundamentally misaligned embeddings in the latent space, so standard contrastive loss fails to reconcile these divergent representations, preventing LVLMs from learning a unified embedding space. To address the gap between two types of inputs and enable consistent cross-modal projection, we introduce an auxiliary loss tailored for multi-modal inputs.
Given input $q$ (or $c_i$) containing both image and text modalities,
we construct $q'$ (or $c_i^{'}$) by removing the image component and generating pseudo visual tokens through our completion module. 
\begin{equation}
    \mathcal{L}_2 = \frac{1}{B} \sum_{i=1}^B \left[ \mathcal{H}(E_{c}^{(i)}, E_{c'}^{(i)}) + \mathcal{H}(E_q^{(i)}, E_{q'}^{(i)}) \right],
\end{equation}
where $\mathcal{H}(\cdot,\cdot)$ denotes the cross-entropy function.
This proposed loss objective naturally accommodates uni-modal cases:
when the input $q$ lacks visual content, $q'$ becomes identical to $q$, nullifying their cross-entropy contribution.

The composite loss function combines these objectives through linear combination:
\begin{equation}
    \mathcal{L} = \mathcal{L}_1 + \alpha\mathcal{L}_2,
\end{equation}
where $\alpha \in \mathbb{R}^+$ controls the relative importance of cross-modal alignment.
Our framework jointly optimizes two objectives:
discriminative embedding learning $\mathcal{L}_1$ and modality-invariant representation learning via $\mathcal{L}_2$.
For complete modalities, $\mathcal{L}_1$ reduces distances between positive pairs while separating negative pairs.
When modalities are missing, $\mathcal{L}_2$ bridges pseudo image embeddings with real counterparts.
The orthogonal combination of them constructs a unified embedding space that preserves discriminative power across modalities and robustness to missing-modality data.


\section{Evaluation}\label{sec:evaluation}

\subsection{Setup}

In our study, we employ Phi-3.5V~\cite{abdin2024phi3technicalreporthighly} and Qwen2-VL-7B~\cite{wang2024qwen2} as foundational LVLMs, while utilizing Phi-1.5 and Qwen2-1.5B as their T2I counterparts. The loss function incorporates a temperature parameter of 0.02 and a hyperparameter $\alpha$ of 0.1. For Phi-3.5V based models, we process each image into 4 sub-image crops, whereas for Qwen2-VL-7B based models, all images are resized to a standardized resolution of 672$\times$672. For details of experimental settings, please refer to Appendix~\ref{sec:setting}.

For training data, we utilize the MMEB~\cite{jiang2024vlm2vec}, comprising 20 diverse datasets across four domains: classification, retrieval, VQA, and visual grounding. By subsampling datasets exceeding 50K instances into 50K samples, we construct a final training set of 662K samples.

The evaluation framework also builds upon MMEB, incorporating both in-distribution (20 datasets) and out-of-distribution (16 datasets) test sets, with each dataset containing 1K samples. We employ Precision@1 as the primary metric for assessing model performance across all benchmarks.

For evaluation baselines, we employ several multi-modal embedding models, including CLIP~\cite{radford2021learning}, OpenCLIP~\cite{cherti2023reproducible}, BLIP2~\cite{li2023blip}, and SigLIP~\cite{zhai2023sigmoid}, all of which utilize vision or language encoders to generate feature representations. Additionally, we incorporate UniIR~\cite{wei2024uniir} and VLM2VEC~\cite{jiang2024vlm2vec}~\footnote{Here we only include Phi-3.5V variant of VLM2VEC. The rationale for this selection can be found in Appendix~\ref{sec:main_exp_per_task}.}, two models specifically designed for multi-modal embedding tasks. It should be noted that our comparison focuses solely on architectural differences, excluding methods that optimize embeddings from other perspectives~\cite{lan2025llave, yu2025cafe, zhou2024megapairs, chen2025mme5, liu2024lamra, gu2025breaking}. 
These approaches are orthogonal to our work and could complement our architecture in the future.


\subsection{Main Results}

\begin{table*}[tb!]
\centering
\caption{Evaluation results on the MMEB, displaying average meta-task scores. Baseline comparisons include both fine-tuned (FT) and non-FT variants on training data, alongside our Phi-3.5V and Qwen2-VL-7B variants. The notation $(\mathrm{T}+\mathrm{I},\mathrm{T})$ stands for datasets containing multi-modal queries with text targets,. IND/OOD distinguishes between in-distribution and out-of-distribution datasets. The optimal results highlighted in \textbf{bold} and the strongest baseline performances (both FT and non-FT variants) are indicated with \underline{underlines}.}
\label{tab:main_exp}
\resizebox{1.0\linewidth}{!}{
\begin{tabular}{lcccccccccc}
\toprule
\multirow{2}{*}{\textbf{Models}} & \multicolumn{4}{c}{\textbf{Per Meta-Task Score}}  & \multicolumn{6}{c}{\textbf{Average Score}} \\ 
\cmidrule(lr){2-5} \cmidrule(lr){6-11}
                       & Classification & VQA  & Retrieval & Grounding & $(\mathrm{T}+\mathrm{I},\mathrm{T})$ & $(\mathrm{T},\mathrm{T}+\mathrm{I})$ & $(\mathrm{T}+\mathrm{I},\mathrm{T}+\mathrm{I})$ & IND & OOD & Overall \\ \midrule
\# of Datasets $\rightarrow$ & 10 & 10 & 12 & 4 & 22 & 6 & 8 & 20 & 16 & 36 \\ \midrule

\multicolumn{11}{c}{\textbf{\emph{w/o Fine-tuning on MMEB}}} \\ \midrule
CLIP~\citep{radford2021learning}   & 42.8 & 9.1 &  53.0 &  51.8 & 29.8 & 62.1 & 41.6 &  37.1  &  38.7 &  37.8 \\
OpenCLIP~\citep{cherti2023reproducible}  & \underline{47.8}  &  10.9 & 52.3  & 53.3 & \underline{33.1} & 57.9 & 44.2 &  39.3 &  40.2 & 39.7 \\
BLIP2~\citep{li2023blip}  & 27.0  &  4.2 & 33.9  & 47.0 & 15.7 & 43.1 & 37.9 &  25.3 &  25.1 & 25.2 \\
SigLIP~\citep{zhai2023sigmoid}  & 40.3  &  8.4 & 31.6  & 59.5 & 27.0 & 44.6 & 49.0 &  32.3 &  38.0 & 34.8 \\ 
UniIR~\cite{wei2024uniir} & 42.1 & \underline{15.0} & \underline{60.1} & \underline{62.2} & 32.5 & \underline{58.2} & \underline{59.7} & \underline{44.7} & \underline{40.4} & \underline{42.8} \\ 
\rowcolor{yellow!15}
$\Delta$ w/o fine-tune & $\uparrow$12.8  & $\uparrow$43.2 &$\uparrow$4.9 &$\uparrow$20.2& $\uparrow$26.5 & $\uparrow$15.4 &$\uparrow$5.4 &$\uparrow$23.5 & $\uparrow$18.0 &$\uparrow$20.4 \\ 
\midrule

\multicolumn{11}{c}{\textbf{\emph{w/ Fine-tuning on MMEB}}} \\ \midrule
CLIP-FT & 50.0 & 27.0 & 55.3 & 64.8 & 40.3 & 64.7 & 49.3 & 52.2 & 38.9 & 47.0 \\
OpenCLIP-FT & 51.4 & 30.8 & 58.1 & 66.3 & 43.1 & 66.8 & 54.4 & 56.9 & 40.4 & 49.6 \\
VLM2VEC (Phi-3.5V)~\cite{jiang2024vlm2vec} & \underline{54.8} & \underline{54.9} & \underline{62.3} & \underline{79.5} & \underline{56.1} & \underline{72.6} & \underline{61.5} & \underline{66.5} & \underline{52.0} & \underline{60.1} \\ 
\rowcolor{gray!10}
\textbf{UniMoCo (Phi-3.5V)} & 55.0 & \textbf{58.2} & 63.2  & \textbf{82.4} & 57.7 & 72.8 & 64.1 & \textbf{68.2} & 53.5 & 61.7 \\ 
\rowcolor{gray!10}
\textbf{UniMoCo (Qwen2-VL-7B)} & \textbf{62.6} & 55.5 & \textbf{65.0} & 78.2 & \textbf{59.6} & \textbf{73.6} & \textbf{65.1} & 67.0 & \textbf{58.4} & \textbf{63.2} \\ 
\rowcolor{yellow!15}
$\Delta$ w/ fine-tune & $\uparrow$7.8& $\uparrow$3.3 & $\uparrow$2.7 & $\uparrow$2.9 & $\uparrow$3.5 & $\uparrow$1.0 & $\uparrow$3.6  & $\uparrow$1.7 & $\uparrow$6.4 & $\uparrow$3.1 \\ 
\bottomrule
\end{tabular}
}
\end{table*}

The results presented in Table~\ref{tab:main_exp} demonstrate that our proposed UniMoCo framework outperforms all baselines on the MMEB. Detailed results appear in Appendix~\ref{sec:main_exp_per_task}. Notably, the Qwen2-VL variant achieves the best overall performance with a score of 63.2, comprising 67.0 on in-distribution datasets and 58.4 on out-of-distribution datasets. When examining performance across different modality combinations, this variant attains scores of 59.6, 73.6, and 65.1 for $(\mathrm{T}+\mathrm{I},\mathrm{T})$, $(\mathrm{T},\mathrm{T}+\mathrm{I})$, and $(\mathrm{T}+\mathrm{I},\mathrm{T}+\mathrm{I})$ tasks respectively.

Our analysis reveals significant improvements over existing approaches. Compared to the strongest baseline without fine-tuning, UniMoCo shows substantial gains of 12.8, 43.2, 4.9, and 20.2 points on classification, VQA, retrieval, and grounding meta-tasks, respectively. Compared to fine-tuned baselines, our method maintains consistent improvements of 7.8, 3.3, 2.7, and 2.9 points.

We observe that $(\mathrm{T}+\mathrm{I},\mathrm{T})$ tasks dominates the majority of the datasets in the MMEB training set (13 out of 20). A similar trend emerges when comparing Phi-3.5V based VLMVEC and our UniMoCo: while VLMVEC performs comparably to UniMoCo on $(\mathrm{T}+\mathrm{I},\mathrm{T})$ tasks, our method significantly outperforms it on other modality combinations. This reveals that the failure of traditional architectures to properly align all modality completions results in a bias toward frequent training combinations. In comparison, UniMoCo overcomes this challenge by unifying all modalities within one aligned architecture, resulting in uniformly better performance across all combinations. Additional analysis can be found in Section~\ref{sec:results_analysis}.

\subsection{Modality Combination Bias Analysis}\label{sec:results_analysis}

This study investigates whether traditional model architectures exhibit inherent biases toward specific modality combinations and evaluates the potential of our proposed UniMoCo framework in addressing this limitation. To ensure a fair comparison, we conduct extensive experiments using both VLM2VEC and our UniMoCo approach, with Phi-3.5V serving as the backbone LVLM. For our experimental setup, we construct a specialized training set from MMEB's retrieval-related classes, comprising three distinct subsets corresponding to different modality combinations. To control data distribution, we create three training set variants, each with half the samples from one modality combination and the rest equally split between the other two. Models trained on these imbalanced sets are evaluated on MMEB, thus enabling rigorous examination of modality preferences.

\begin{figure*}
    \centering
    \includegraphics[width=1\linewidth]{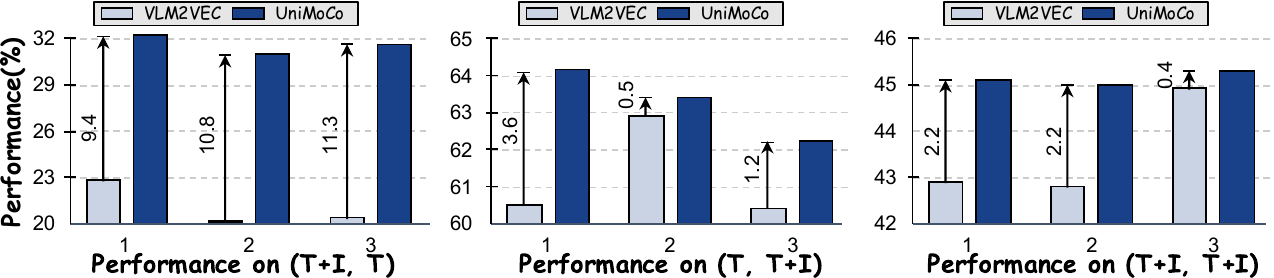}
    \caption{Performance of models trained on skewed training data distributions. The x-axis labels (1, 2, 3) correspond to training sets dominated by $(\mathrm{T}+\mathrm{I},\mathrm{T})$, $(\mathrm{T},\mathrm{T}+\mathrm{I})$, and $(\mathrm{T}+\mathrm{I},\mathrm{T}+\mathrm{I})$ combinations respectively.}
    \label{fig:result_analysis}
\end{figure*}

As demonstrated in Figure~\ref{fig:result_analysis}, traditional approaches are indeed susceptible to imbalanced modality combinations in the training data. VLM2VEC performs strongly when evaluated on tasks matching its dominant training modality combination. For instance, when trained on $(\mathrm{T},\mathrm{T}+\mathrm{I})$ dominated datasets, VLM2VEC achieves a score of 62.9, approaching UniMoCo's performance of 63.4. However, its performance significantly deteriorates on other combinations (42.8 vs. 45.0 and 20.1 vs. 30.9 for UniMoCo). Furthermore, VLM2VEC trained on other modality distributions show markedly reduced performance on $(\mathrm{T},\mathrm{T}+\mathrm{I})$ evaluation tasks, scoring only 60.5 and 61.0, respectively. This limitation remains consistent across all evaluations examined.

In contrast, our UniMoCo framework demonstrates remarkable robustness, maintaining consistently high performance across all evaluation benchmarks regardless of the dominant modality combination in the training data. These results clearly indicate that UniMoCo effectively addresses the critical limitation of modality bias.

\subsection{Ablation Study}\label{sec:ablation}

\begin{table}
\centering
\caption{Evaluation results on MMEB across different architectural designs. Baseline represents VLM2VEC; other configurations show UniMoCo variants.}
\label{tab:ablation}
\resizebox{\columnwidth}{!}{
\begin{tabular}{lcccccc}
\toprule
\multirow{2}{*}{\textbf{Models}} & \multicolumn{6}{c}{\textbf{Average Score}} \\ 
\cmidrule(lr){2-7}
                       & $(\mathrm{T}+\mathrm{I},\mathrm{T}+\mathrm{I})$ & $(\mathrm{T}+\mathrm{I},\mathrm{T})$ & $(\mathrm{T},\mathrm{T}+\mathrm{I})$ & IND & OOD & Overall \\ \midrule
\# of Datasets $\rightarrow$ & 22 & 6 & 8 & 20 & 16 & 36 \\ \midrule

Baseline & 52.9 & 68.4 & 60.0 & 62.4 & 50.1 & 56.9 \\ 
T2I Only & 49.6 & 64.7 & 57.2 & 58.9 & 48.4 & 53.6 \\ 
w/ Encoder & 51.9 & 66.5 & 57.6 & 61.2 & 48.1 & 55.4 \\
w/ Padding & 53.3 & 69.5 & \textbf{60.5} & 63.5 & 49.8 & 57.4 \\
w/ Encoder + Padding & \textbf{54.0} & \textbf{70.1} & 60.5 & \textbf{64.3} & 50.1 & \textbf{58.0} \\

\bottomrule
\end{tabular}
}
\end{table}

\subsubsection{Modality-Completion Module}
To thoroughly examine the design choices for our modality-completion module, we compare various architectural variants including configurations with and without padding tokens as well as additional vision encoders as proposed in Section~\ref{sec:arch}.

Table~\ref{tab:ablation} reveals that simply using a single T2I model without proper modifications leads to significant performance degradation. This occurs because the T2I model produces output embeddings with lengths corresponding to the input query text. In typical scenarios where query texts rarely exceed 40 tokens, this produces short pseudo visual embeddings that must be matched against real image embeddings containing at least 576 patch tokens. Such significant length discrepancy creates challenges for computing meaningful similarity between these representations. Our padding strategy using optimized prompts and dummy tokens to align the T2I model's input length with real image token counts can generate more representative pseudo visual embeddings. As shown in Table~\ref{tab:ablation}, this approach consistently enhances performance across all modality combinations by better leveraging the T2I model's latent capabilities.

Since pseudo visual embeddings still differ from their real counterparts, employing a dedicated vision encoder to process these embeddings yields additional gains (53.6 to 55.4). However, this gain is outweighed by padding (55.4 vs. 57.4) due to the persistent issue of dimensional mismatch within the T2I processing pipeline.

Furthermore, the synergistic combination of both techniques yields superior improvements over the baseline, particularly for the $(\mathrm{T}+\mathrm{I},\mathrm{T}+\mathrm{I})$ and $(\mathrm{T},\mathrm{T}+\mathrm{I})$ tasks that heavily depend on robust modality completion. These results validate our approach's capability to effectively align diverse modality combinations with robust performance.

\begin{table}[tb!]
    \centering
    \caption{Performance evaluation across varying T2I model scales on MMEB, demonstrating how model scale impacts multi-modal representation learning.}
    \label{tab:t2i_model}
    \resizebox{\columnwidth}{!}{
    \begin{tabular}{lcccccc}
        \toprule
        \multirow{2}{*}{\textbf{T2I Model Size}} & \multicolumn{6}{c}{\textbf{Average Score}} \\ 
        \cmidrule(lr){2-7}
        & $(\mathrm{T}+\mathrm{I},\mathrm{T})$ & $(\mathrm{T},\mathrm{T}+\mathrm{I})$ & $(\mathrm{T}+\mathrm{I},\mathrm{T}+\mathrm{I})$ & IND & OOD & Overall \\ \midrule
        \# of Datasets $\rightarrow$ & 22 & 6 & 8 & 20 & 16 & 36 \\ \midrule

        0.5B & 47.5 & 68.4 & 53.4 & 58.9 & 44.0  & 52.3 \\ 
        1.5B & 54.0 & 70.1 & 60.5 & 64.3 & 50.1 & 58.0  \\ 
        7B & \textbf{56.3} & \textbf{73.3} & \textbf{62.2} & \textbf{67.2}  & \textbf{52.0} & \textbf{60.4} \\

        \bottomrule
    \end{tabular}
    }
\end{table}

\subsubsection{T2I Model}
Table~\ref{tab:t2i_model} investigates how model scale affects multi-modal embedding performance using Qwen2-VL-7B as the LVLM backbone with three language model variants (Qwen2-0.5B, Qwen2-1.5B, Qwen2-7B). Our experiments demonstrate a clear relationship between model size and task performance, showing that larger T2I models consistently improve UniMoCo’s embedding quality in all benchmarks and modality combinations. This phenomenon aligns with established scaling laws~\cite{jia2021scaling}, as increased model capacity improves domain adaptation and semantic representation capabilities critical for generating high-fidelity pseudo visual embeddings from textual inputs. The 7B variant demonstrates significantly improved performance by achieving greater accuracy in text-to-embedding translation, especially in tasks that demand precise latent space mapping. These findings underscore the importance of model scale in multi-modal systems, demonstrating that parameter expansion in specialized components can significantly boost effectiveness.

\begin{table}
\centering
\caption{Performance evaluation showing the impact of $\alpha$, which balances contrastive loss $\mathcal{L}_1$ and auxiliary loss $\mathcal{L}_2$ in our objective function.}
\label{tab:loss}
\resizebox{\columnwidth}{!}{
\begin{tabular}{lcccccc}
\toprule
\multirow{2}{*}{\textbf{Configurations}} & \multicolumn{6}{c}{\textbf{Average Score}} \\ 
\cmidrule(lr){2-7}
                      & $(\mathrm{T}+\mathrm{I},\mathrm{T})$ & $(\mathrm{T},\mathrm{T}+\mathrm{I})$ & $(\mathrm{T}+\mathrm{I},\mathrm{T}+\mathrm{I})$ & IND & OOD & Overall \\ \midrule
\# of Datasets $\rightarrow$ & 22 & 6 & 8 & 20 & 16 & 36 \\ \midrule

$\alpha=0.0$ & 54.0 & 70.1 & 60.5 & 64.4 & 50.0 & 58.0 \\ 
$\alpha=0.1$ & 54.6 & 71.1 & 60.5 & 64.5 & 51.1 & 58.5 \\ 
$\alpha=0.2$ & \textbf{55.3} & \textbf{71.8} & \textbf{61.1} & \textbf{64.8} & \textbf{52.1} & \textbf{59.2}\\
$\alpha=0.3$ & 55.2 & 71.0 & 60.6 & 65.0 & 51.3 & 58.9 \\
$\alpha=0.4$ & 54.5 & 71.6 & 60.5 & 64.7 & 50.8 & 58.5 \\

\bottomrule
\end{tabular}
}
\end{table}

\subsubsection{Training Strategy}
As demonstrated in Table~\ref{tab:loss}, the integration of auxiliary loss for multi-modal inputs significantly enhances model performance from all aspects. It simultaneously enhances downstream task performance, improves generalization on OOD, and strengthens robustness against diverse modality combinations.
Our experiments show consistent performance gains when increasing the hyperparameter $\alpha$ from 0.0 to 0.2. We attribute it to the auxiliary loss's role in accelerating the convergence of the modality-completion module, thereby generating higher-quality pseudo visual embeddings that facilitate better query-target matching. However, further increasing $\alpha$ beyond 0.2 to 0.4 results in performance degradation. This suggests that excessive weighting of the auxiliary loss may shift the primary training objective away from the target of learning effective multi-modal embeddings. Through comprehensive evaluation, we identify $\alpha=0.2$ as the optimal balance point that maintains the primary training objective while preserving enhanced transfer capabilities.

Notably, the most significant improvements occur in $(\mathrm{T}+\mathrm{I},\mathrm{T})$ and $(\mathrm{T},\mathrm{T}+\mathrm{I})$ datasets, which aligns with our  design since these scenarios involve missing image modality where the auxiliary loss practically bridges the gap between generated pseudo visual embeddings and real visual embeddings. The observed improvement in $(\mathrm{T}+\mathrm{I},\mathrm{T}+\mathrm{I})$ configurations may stem from the auxiliary loss's secondary benefit of aligning embedding spaces between different vision encoders. This alignment enables language models to process inputs from more consistent embedding space, thereby improving their processing ability. The consistent performance gains across different modality combinations validate that appropriate auxiliary supervision can simultaneously enhance both modality completion and cross-modal alignment.

\subsubsection{Efficiency Analysis}

To comprehensively evaluate our framework, we quantify the overhead of Phi-3.5V based UniMoCo against the baseline (w/o the completion module and $\mathcal{L}_2$ loss) under the same backbone.

During training, UniMoCo requires approximately 35\% more latency and consumes 20\% higher peak memory. In inference, however, the overhead drops significantly to just 8\% added latency and 10\% memory increase. The cost disparity between training and inference arises because $\mathcal{L}_2$ loss generates more activations during training, consuming more memory and latency, whereas inference involves no loss computation and thus incurs less overhead. We argue that this moderate and acceptable increase in computational resources, particularly the minor overhead at inference, is a justifiable trade-off for UniMoCo's significant and consistent performance gains.

\section{Conclusion}

In this paper, we propose UniMoCo which incoraprates LVLM with modality completion effectively handles diverse modality combinations in embedding tasks. It outperforms existing methods and reduces bias caused by modality imbalance in training data. This approach ensures robust and consistent performance across various scenarios.

\section*{Limitation}\label{sec:limitation}

Our study addresses modality combination bias but still leaves several areas for future work. We focus on structural innovations without fully exploring other approaches like training data enhancement or contrastive learning optimization. Combining these with our architecture could produce more robust multi-modal representations. Furthermore, our experiments primarily use the MMEB, but testing on diverse benchmarks would better validate generalizability. These potential extensions suggest a clear pathway for advancing UniMoCo's performance.


{
\small
\bibliography{ref/reference, ref/Top}
}

\clearpage
\appendix
\section{Details of Evaluation Settings}\label{sec:setting}

\subsection{Evaluation Datasets}

For the main experiments, UniMoCo is trained on 20 diverse datasets encompassing multiple tasks: classification (ImageNet-1K, N24News, HatefulMemes, VOC2007, SUN397), VQA (OK-VQA, A-OKVQA, DocVQA, InfographicsVQA, ChartQA, Visual7, ScienceQA), (VisDial, CIRR, VisualNews\_i2t, VisualNews\_t2i, MSCOCO\_t2i, MSCOCO\_i2t, NIGHTS, WebQA), and visual grounding (MSCOCO). Evaluation is conducted on 36 test datasets as specified in Table~\ref{tab:main_exp_per_task}, with all datasets sourced from MMEB~\cite{jiang2024vlm2vec}.

For additional experiments, we maintain the same training configuration except in the modality combination bias analysis. In this specific evaluation, we select three representative base datasets: VisualNews\_i2t $(\mathrm{T}+\mathrm{I},\mathrm{T})$, VisDial $(\mathrm{T},\mathrm{T}+\mathrm{I})$, and NIGHTS $(\mathrm{T}+\mathrm{I},\mathrm{T}+\mathrm{I})$. We construct three distinct training sets by varying the sample distributions: (1) 10,000 samples from VisualNews\_i2t with 5,000 each from VisDial and NIGHTS; (2) 10,000 from VisDial with 5,000 each from VisualNews\_i2t and NIGHTS; (3) 10,000 from NIGHTS with 5,000 each from VisDial and VisualNews\_i2t. Both our approach and the VLM2VEC baseline are trained on these configurations, with evaluation performed across all 36 MMEB datasets to produce the results discussed in Section~\ref{sec:results_analysis}.

Regarding modality combinations, the datasets can be categorized into three types based on their input-output configurations: The $(\mathrm{T}+\mathrm{I},\mathrm{T})$ group includes ImageNet-1K, N24News, HatefulMemes, VOC2007, SUN397, Place365, ImageNet-A, ImageNet-R, ObjectNet, Country211, VisualNews\_i2t, MSCOCO\_i2t, OK-VQA, A-OKVQA, DocVQA, InfographicsVQA, ChartQA, Visual7, ScienceQA, VizWiz, GQA, TextVQA. The $(\mathrm{T},\mathrm{T}+\mathrm{I})$  category comprises VisDial, VisualNews\_t2i, MSCOCO\_t2i, WebQA, Wiki-SS-NQ, EDIS. Lastly,  the $(\mathrm{T}+\mathrm{I},\mathrm{T}+\mathrm{I})$ combination contains CIRR, NIGHTS, FashionIQ, OVEN, MSCOCO, RefCOCO, RefCOCO-Matching, Visual7W-Pointing. Among the training datasets, 13 belong to $(\mathrm{T}+\mathrm{I},\mathrm{T})$, 3 to $(\mathrm{T},\mathrm{T}+\mathrm{I})$, and 4 to $(\mathrm{T}+\mathrm{I},\mathrm{T}+\mathrm{I})$. The predominance of $(\mathrm{T}+\mathrm{I},\mathrm{T})$ datasets in the training set introduces a bias, as discussed in Section~\ref{sec:results_analysis}.

\subsection{Hyperparameters and Computational Requirements}\label{sec:paramater}

Table~\ref{tab:supp_training_parameters} presents our detailed setting during training and test.

\section{Extended Benchmark on Flickr30K}\label{sec:flickr30k}

To address concerns about generalization beyond MMEB and demonstrate robustness in real-world scenarios, we conduct an additional evaluation on Flickr30K~\cite{plummer2015flickr30k}, a widely-adopted zero-shot text-image retrieval benchmark featuring natural images with human-annotated captions. Unlike MMEB's curated task-specific datasets, Flickr30K presents challenges from diverse visual content and naturally occurring linguistic variations in user-generated descriptions, providing a complementary testbed for assessing model generalization. 

We compare \model against established baselines including CLIP-L/14 and VLM2VEC, with baseline results sourced from~\cite{jiang2024vlm2vec}. Table~\ref{tab:zero-shot-flickr30k} summarizes the R@{1,5,10} scores for both text-to-image and image-to-text retrieval directions. Results demonstrate that \model consistently outperforms all competitors across all metrics, achieving substantial improvements over VLM2VEC (e.g., +2.8 R@1 for image retrieval and +2.2 R@1 for text retrieval), thereby confirming the effectiveness of our modality completion approach beyond the MMEB evaluation framework.

\begin{table*}[!ht]
  \small
    \centering
    \caption{Zero-shot text-image retrieval performance on Flickr30K. Baseline results (CLIP-L/14 and VLM2VEC) are sourced from~\cite{jiang2024vlm2vec}. \model (Qwen2-VL-7B based) achieves uniformly higher R@{1,5,10} in both text-to-image and image-to-text retrieval, demonstrating the effectiveness of modality completion beyond MMEB.}
    \begin{tabular}{lcccccc}
        \toprule
        \textbf{Model} & \multicolumn{3}{c}{\textbf{image} retrieval} & \multicolumn{3}{c}{\textbf{text} retrieval} \\
        \cmidrule(lr){2-4} \cmidrule(lr){5-7}
        & R@1 & R@5 & R@10 & R@1 & R@5 & R@10  \\
        \midrule
        CLIP-L/14        &  65.2  & 87.3 &  92.0 & 85.2 & 97.3 & 99.0  \\
        VLM2VEC          &  80.3  & 95.0 &  97.4 & 94.6 & 99.5 & 99.8  \\
        \model           &  \textbf{83.1} & \textbf{96.2} & \textbf{98.1} & \textbf{96.8} & \textbf{99.7} & \textbf{99.9} \\
        \bottomrule
    \end{tabular}
    \label{tab:zero-shot-flickr30k}
\end{table*}

\begin{table*}[h!]
\centering
\caption{Hyperparameters and computational requirements for UniMoCo (Phi-3.5V) and UniMoCo (Qwen2-VL-7B) during training and test.}
\label{tab:training_details}
\begin{tabular}{@{}lcc@{}}
\toprule
\multicolumn{1}{c}{\textbf{Hyperparameter}} & \textbf{UniMoCo (Phi-3.5V)} & \textbf{UniMoCo (Qwen2-VL-7B)} \\
\midrule
\multicolumn{3}{c}{\textbf{Training Setting}} \\ \midrule
Resolution & $336\times 336$ & $672\times 672$ \\
Training samples & \multicolumn{2}{c}{662K} \\
Number of Samples per Dataset &  \multicolumn{2}{c}{50K} \\
Batch size & \multicolumn{2}{c}{1024} \\
Learning rate & 6$\times$10$^{-5}$ & 1$\times$10$^{-4}$\\
LoRA rank & \multicolumn{2}{c}{8} \\
Steps & \multicolumn{2}{c}{2K} \\
GPU configuration & \multicolumn{2}{c}{8$\times$A100} \\
Precision & \multicolumn{2}{c}{BF16} \\
Training time & \textasciitilde135 hours & \textasciitilde185 hours \\ \midrule
\multicolumn{3}{c}{\textbf{Test Setting}} \\ \midrule
Test samples & \multicolumn{2}{c}{36K} \\
Number of Samples per Dataset &  \multicolumn{2}{c}{1K} \\
Batch size & \multicolumn{2}{c}{16} \\
GPU configuration & \multicolumn{2}{c}{1$\times$A100} \\
Precision & \multicolumn{2}{c}{BF16} \\
Test time & \textasciitilde3 hours & \textasciitilde10 hours \\
\bottomrule
\end{tabular}
\label{tab:supp_training_parameters}
\vspace{-3mm}
\end{table*}

\section{Specific Results on MMEB}\label{sec:main_exp_per_task}

Table~\ref{tab:main_exp_per_task} provides comprehensive experimental results corresponding to the data presented in Table~\ref{tab:main_exp}. It is worth noting that VLMVEC~\cite{jiang2024vlm2vec} has also introduced model variants based on LLaVA-1.6 and Qwen2-VL, which were not included in our comparative analysis for the following reasons. Firstly, as documented in their materials, these models were trained using a different dataset configuration, employing 100K samples per dataset compared to our 50K sample size. Secondly, they adopted a higher input resolution of 1344$\times$1344 pixels, which differs from our standardized resolution of 672$\times$672 pixels. Due to these substantial differences in training settings and model configurations, we considered a direct performance comparison would not yield fair or meaningful results, thus justifying their exclusion from our evaluations.

\begin{table*}[]
\centering
\caption{The detailed results of the baselines and our UniMoCo on MMEB. OOD are highlighted with a yellow background in the table. Here UniMoCo-1 uses Phi-3.5V as backbone LVLM while UniMoCo-2 uses Qwen2-VL-7B as backbone LVLM.}
\resizebox{1.0\textwidth}{!}{
\begin{tabular}{lcccccccc}
\toprule
\rowcolor{gray!30}  & \textbf{CLIP} & \textbf{OpenCLIP} & \textbf{SigLIP} & \textbf{BLIP2} & \textbf{UniIR} & \textbf{VLM2VEC} & \textbf{UniMoCo-1} & \textbf{UniMoCo-2} \\
\midrule
\rowcolor{orange!30} \textbf{Classification (10 tasks)} & & & & & & & & \\
ImageNet-1K          & 55.8 & 63.5 & 45.4 & 10.3 & 58.3 & 65.6 & 62.7 & 75.2 \\
N24News              & 34.7 & 38.6 & 13.9 & 36.0 & 42.5 & 79.5 & 81.7 & 69.5 \\
HatefulMemes         & 51.1 & 51.7 & 47.2 & 49.6 & 56.4 & 67.1 & 71.0 & 77.0 \\
VOC2007              & 50.7 & 52.4 & 64.3 & 52.1 & 66.2 & 88.6 & 87.1 & 84.5 \\
SUN397               & 43.4 & 68.8 & 39.6 & 34.5 & 63.2 & 72.7 & 69.7 & 74.1 \\
\rowcolor{yellow!15} Place365  & 28.5 & 37.8 & 20.0 & 21.5 & 36.5 & 42.6 & 42.7 & 44.0 \\
\rowcolor{yellow!15} ImageNet-A & 25.5 & 14.2 & 42.6 & 3.2  & 9.8 & 19.3 & 23.0 & 47.3 \\
\rowcolor{yellow!15} ImageNet-R & 75.6 & 83.0 & 75.0 & 39.7 & 66.2 & 70.2 & 72.2 & 84.1 \\
\rowcolor{yellow!15} ObjectNet  & 43.4 & 51.4 & 40.3 & 20.6 & 32.2& 29.5 & 23.5 & 39.6 \\
\rowcolor{yellow!15} Country-211 & 19.2 & 16.8 & 14.2 & 2.5  & 11.3 & 13.0 & 16 & 29.6 \\
\textit{All Classification} & 42.8 & 47.8 & 40.3 & 27.0 & 44.3 & 54.81 & 55.0 & 62.6 \\
\midrule

\rowcolor{blue!30} \textbf{VQA (10 tasks)} & & & & & & & & \\
OK-VQA               & 7.5  & 11.5 & 2.4  & 8.7  & 25.4 & 63.2 & 65.5 & 65.0 \\
A-OKVQA              & 3.8  & 3.3  & 1.5  & 3.2  & 8.8 & 50.2 & 54.0 & 55.6 \\
DocVQA               & 4.0  & 5.3  & 4.2  & 2.6  & 6.2 & 78.4 & 78.5 & 83.6 \\
InfographicsVQA      & 4.6  & 4.6  & 2.7  & 2.0  & 4.6 & 40.8 & 43.3 & 47.6 \\
ChartQA              & 1.4  & 1.5  & 3.0  & 0.5  & 1.6 & 59.0 & 57.8 & 53.2 \\
Visual7W             & 4.0  & 2.6  & 1.2  & 1.3  & 14.5 & 47.7 & 52.3 & 48 \\
\rowcolor{yellow!15} ScienceQA  & 9.4  & 10.2 & 7.9  & 6.8  & 43.4 & 42.1 & 51.2 & 44.9 \\
\rowcolor{yellow!15} VizWiz    & 8.2  & 6.6  & 2.3  & 4.0  & 24.3 & 39.2 & 40 & 41.4 \\
\rowcolor{yellow!15} GQA        & 41.3 & 52.5 & 57.5 & 9.7  & 48.8 & 60.7 & 69.1 & 42.2 \\
\rowcolor{yellow!15} TextVQA    & 7.0  & 10.9 & 1.0  & 3.3  & 15.1 & 66.1 & 70.5 & 73.4 \\
\textit{All VQA}      & 9.1  & 10.9 & 8.4  & 4.2  & 16.2 & 54.9 & 58.2 & 55.5 \\
\midrule

\rowcolor{green!30} \textbf{Retrieval (12 tasks)} & & & & & & & & \\
VisDial              & 30.7 & 25.4 & 21.5 & 18.0 & 42.2 & 73.3 & 75.5 & 72.6 \\
CIRR                 & 12.6 & 15.4 & 15.1 & 9.8  & 51.3 & 47.8 & 50.0 & 51.0 \\
VisualNews\_t2i      & 78.9 & 74.0 & 51.0 & 48.1 & 74.3 & 67.2 & 68.5 & 73.0 \\
VisualNews\_i2t      & 79.6 & 78.0 & 52.4 & 13.5 & 76.8 & 70.7 & 70.6 & 69.4 \\
MSCOCO\_t2i          & 59.5 & 63.6 & 58.3 & 53.7 & 68.5 & 70.6 & 71.7 & 70.7 \\
MSCOCO\_i2t          & 57.7 & 62.1 & 55.0 & 20.3 & 72.1 & 66.5 & 67.9 & 61.3 \\
NIGHTS               & 60.4 & 66.1 & 62.9 & 56.5 & 66.2 & 66.1 & 67.5 & 67.9 \\
WebQA                & 67.5 & 62.1 & 58.1 & 55.4 & 89.6 & 88.1 & 88.5 & 71.0 \\
\rowcolor{yellow!15} FashionIQ  & 11.4 & 13.8 & 20.1 & 9.3  & 40.2 & 12.9 & 16.1 & 21.2 \\
\rowcolor{yellow!15} Wiki-SS-NQ & 55.0 & 44.6 & 55.1 & 28.7 & 12.2 & 56.6 & 59.5 & 66.9 \\
\rowcolor{yellow!15} OVEN       & 41.1 & 45.0 & 56.0 & 39.5 & 69.4 & 47.3 & 49.3 & 68.1 \\
\rowcolor{yellow!15} EDIS       & 81.0 & 77.5 & 23.6 & 54.4 & 79.2 & 79.9 & 73.2 & 87.3 \\
\textit{All Retrieval} & 53.0 & 52.3 & 31.6 & 33.9 & 61.8 & 62.3 & 63.2 & 65.0 \\
\midrule

\rowcolor{purple!30} \textbf{Visual Grounding (4 tasks)} & & & & & & & & \\
MSCOCO         & 33.8 & 34.5 & 46.4 & 28.9 & 46.6 & 67.3 & 79.7 & 68.5 \\
\rowcolor{yellow!15} RefCOCO  & 56.9 & 54.2 & 70.8 & 47.4 & 67.8 & 84.7 & 85.7 & 83.3 \\
\rowcolor{yellow!15} RefCOCO-matching  & 61.3 & 68.3 & 50.8 & 59.5 & 62.9 & 79.2 & 79.9 & 85.8 \\
\rowcolor{yellow!15} Visual7W-pointing & 55.1 & 56.3 & 70.1 & 52.0 & 71.3 & 86.8 & 84.4 & 75.0 \\
\textit{All Visual Grounding} & 51.8 & 53.3 & 59.5 & 47.0 & 65.3 & 79.5 & 82.4 & 78.2 \\
\midrule

\rowcolor{cyan!15} \textbf{Final Score (36 tasks)} & & & & & & & & \\
All                  & 37.8 & 39.7 & 34.8 & 25.2 & 42.8 & 60.1 & 61.7 & 63.2 \\
All IND       & 37.1 & 39.3 & 32.3 & 25.3 & 47.1 & 66.5 & 68.2 & 67.0 \\
All OOD       & 38.7 & 40.2 & 38.0 & 25.1 & 41.7 & 52.0 & 53.5 & 58.4 \\
All $(\mathrm{T}+\mathrm{I},\mathrm{T})$ & 29.8 & 33.1 & 15.7 & 27.0 & 32.5 & 56.1 & 57.7 & 59.6 \\
All $(\mathrm{T},\mathrm{T}+\mathrm{I})$ & 62.1 & 57.9 & 43.1 & 44.6 & 58.2 & 72.6 & 72.8 & 73.6 \\
All $(\mathrm{T}+\mathrm{I},\mathrm{T}+\mathrm{I})$ & 41.6 & 44.2 & 37.9 & 49.0 & 59.7 & 61.5 & 64.1 & 65.1 \\

\bottomrule
\end{tabular}
}
\label{tab:main_exp_per_task}
\end{table*}

\section{Further Analysis}

\subsection{Design of Padding Tokens}\label{sec:pad}

To ensure consistent input lengths for the modality-completion module, padding tokens are applied to align textual inputs with the fixed number of image tokens (576 tokens for Phi-3.5V-based UniMoCo). The formatted prompt structure follows this template:

\begin{equation}
q_{t}^{'} = [{\mathrm{Padding }\mathrm{Prompt}}] \mathbin\Vert q_t \mathbin\Vert [\mathrm{END}] \mathbin\Vert [\mathrm{dummy}{\mathrm{tokens}}]
\end{equation}

Here, $q_{t}'$ represents the processed input to the modality-completion module, while $q_{t}$ denotes the original textual input. The padding instruction states: \textit{"Image modality is missing in this case. We employ a text-to-image model to generate a highly detailed visual description based on the given instruction and query. The characters following [END] serve as placeholders. Query: "}. The number of dummy tokens after [END] is calculated as $N = 576 - \ell(\mathrm{P}_{\mathrm{pad}}) - \ell(q_t) - 1$ padding, ensuring the total length of $q_{t}'$ matches the required 576 tokens. This padding mechanism maintains structural consistency between textual and visual inputs during processing.

We investigated various approaches for padding token formatting in our experiments. First, we designed task-specific padding prompts where each category (Classification, Retrieval, VQA, or Grounding) received distinct prompts structured as $[\mathrm{Label}][{\mathrm{Padding }\mathrm{Prompt}}]$ where label corresponds to the category name, denoted as Pad-1 in Table~\ref{tab:pad}. Second, we examined varying padding lengths across task categories, with classification tasks padded to half the standard 576 tokens while other tasks retained full padding (Pad-2). This adjustment was based on the observation that classification targets typically involve concise labels (one or two words) that might benefit from shorter pseudo visual embeddings. However, both methods showed minimal performance improvements and sometimes caused degradation, prompting their abandonment.

We further explored padding length variations, testing configurations including half-length padding (288 tokens, denoted as Pad-3). This adjustment demonstrated contrasting effects across different models: while it adversely affected the performance of Phi-3.5V based UniMoCo, it proved beneficial for the Qwen2-VL-7B based implementation. The latter model, operating at $672\times672$ resolution, typically requires $576\times4=2304$ image tokens, but the half-length strategy effectively reduced this requirement to 1152 tokens, resulting in improved efficiency and performance.

Given the substantial variability in input text lengths—ranging from brief single-word labels to comprehensive descriptive responses—we introduced a threshold-based discrimination system (Pad-4) to dynamically adjust padding. Inputs shorter than the predetermined threshold received half-padding, whereas longer inputs retained full-length padding. However, after extensive evaluation across multiple threshold values, we observed negligible performance differences, ultimately leading to the abandonment of this adaptive padding approach.

In light of these findings, we evaluated computational overhead. Although long padding tokens increase latency—as language model processing time scales with sequence length—the overall impact of padding is modest. This stems from: (1) the modality-completion module being substantially smaller than the main LVLM backbone, and (2) the LVLM already handling lengthy sequences of visual, pseudo, and text tokens. Consequently, we can pay less attention to the computational overhead of padding and our padding length selection prioritized model performance over minimal computational savings.

\begin{table*}
\centering
\caption{Performance Comparison of Different Padding Strategies. For Pad-2 and Pad-3, we also tested alternative lengths such as quarter padding. In Pad-4, the token threshold is set to 40. While we experimented with other configurations for Pad-2/3/4, the current settings yielded the best results.}
\label{tab:pad}
\resizebox{0.9\textwidth}{!}{
\begin{tabular}{lccccc}
\toprule
\textbf{Methods} & \textbf{Classification} & \textbf{VQA} & \textbf{Retrieval} & \textbf{Grounding} & \textbf{Overall} \\ 
\midrule
\multicolumn{6}{c}{\textbf{Task-Specific Padding Formats}} \\ 
\midrule
Pad-1 (Category prompts) & 51.5 & 54.3 & 58.9 & 79.9 & 57.9 \\
Pad-2 (Variable lengths) & 52.0 & 52.9 & 59.4 & 81.3 & 58.0 \\  
\midrule
\multicolumn{6}{c}{\textbf{Padding Length Variations}} \\ 
\midrule
Pad-3 (Half-length) & 51.6 & 54.9 & 58.8 & \textbf{{82.5}} & 58.3 \\ 
Pad-4 (Length-adaptive) & 52.2 & 54.6 & \textbf{{60.7}} & 80.6 & 58.9 \\ 
\midrule
\multicolumn{6}{c}{\textbf{Baseline Configuration}} \\ 
\midrule
Standard Padding & \textbf{52.7} & \textbf{55.4} & 60.5 & 80.8 & \textbf{59.2} \\
\bottomrule
\end{tabular}
}
\end{table*}
\subsection{Choice of LoRA Rank}

To reduce computational costs and training time, we employ LoRA for efficient fine-tuning of the models. We conduct experiments with different LoRA ranks using Phi-3.5V as the backbone LVLM. As demonstrated in Table~\ref{tab:lora}, a rank of 8 achieves optimal performance across all tasks. Consequently, we adopt this configuration for subsequent evaluations, including the main results and ablation studies.

\begin{table*}
\centering
\caption{Performance comparison of different LoRA ranks.}
\label{tab:lora}
\resizebox{0.85\textwidth}{!}{
\begin{tabular}{lccccc}
\toprule
\textbf{Methods} & \textbf{Classification} & \textbf{VQA} & \textbf{Retrieval} & \textbf{Grounding} & \textbf{Overall} \\ \midrule
$r=4$ & 52.3 & 51.9 & 56.6 & 76.3 & 56.3 \\
$r=8$ & 52.7 & \textbf{55.4} & \textbf{60.5} & \textbf{80.8} & \textbf{59.2} \\
$r=16$ & \textbf{53.3} & 54.2 & 59.0 & 77.7 & 58.2 \\ 
$r=32$ & 53.3 & 55.4 & 58.5 & 75.4 & 58.1 \\ 
\bottomrule
\end{tabular}
}
\end{table*}

\subsection{Text-missing Modality Analysis}\label{sec:text_missing}

In our design, we incorporate task-specific instructions (e.g., “Represent the given image for classification”) to supply the textual content, thereby focusing primarily on visual-missing (V-missing) scenarios. This strategy follows VLM2VEC~\cite{jiang2024vlm2vec}, whose Table 4 reports performance gains exceeding 20\% for CLIP and VLM2VEC. Furthermore, in real-world applications, system prompts are typically provided when employing LLMs or VLMs, making text-missing (T-missing) cases extremely uncommon. Therefore, we adopt this design and direct our main focus toward V-missing scenarios.

We acknowledge that the current architecture could be extended into a symmetric architecture, where T-missing and V-missing cases are handled uniformly via a modality-completion module that generates pseudo embeddings. Exploring such adaptations is reserved for future work.

\subsection{Uni-modal Performance Analysis}\label{sec:unimodal}

Our benchmark does not include $(\mathrm{T},\mathrm{T})$ tasks, as multi-modal embedding models—including UniMoCo and prior works~\cite{jiang2024vlm2vec,lan2025llave,liu2024lamra,gu2025breaking}—are generally designed for scenarios where images appear in either the query or target. For purely textual retrieval tasks, uni-modal text embedding models remain more appropriate.

Although UniMoCo is a multi-modal framework, it can also operate under uni-modal conditions. We evaluated its text-only performance on the Massive Text Embedding Benchmark (MTEB)~\cite{muennighoff2022mteb}, comparing it with LLM2VEC~\cite{behnamghader2024llm2vec}, an LLM-based text embedding model. The results indicate that UniMoCo achieves competitive performance, with a performance gap of no more than 8\% across tasks. Since UniMoCo is optimized for multi-modal retrieval, and its use of $\mathcal{L}_1$ and $\mathcal{L}_2$ loss may limit its effectiveness in purely textual tasks.





\subsection{Discussions on Components of Modality-Completion Module}\label{sec:ablation_components}
During our evaluations, we find that both the extra vision encoder and the padding mechanism are necessary whenever the visual modality is missing (e.g., in $(\mathrm{T}, \mathrm{T}+\mathrm{I})$ and $(\mathrm{T}+\mathrm{I}, \mathrm{T})$ settings), because the modality-completion module must produce pseudo-visual embeddings that are both length-compatible and distributionally aligned with real visual tokens. Consistent with this observation, the ablation results in Table~\ref{tab:ablation} show that removing either component yields systematic performance drops relative to the full UniMoCo design.

The padding mechanism mainly addresses structural mismatch. Without padding, the T2I module generates pseudo embeddings whose token length is tied to the input text length, which is typically much shorter than the visual token sequence expected by the downstream projector/backbone; this mismatch reduces the effective input volume available to the projector and degrades similarity computation against real-image representations. In contrast, the extra vision encoder targets distribution shift: pseudo embeddings produced in the T2I latent space do not naturally follow the same feature distribution as the LVLM vision encoder outputs, so an additional encoder is required to map them into the backbone's visual representation space. Notably, these issues remain even when the auxiliary loss ($\mathcal{L}_2$) is removed, suggesting that both components are structurally necessary for robust modality completion; therefore, whenever the visual modality is missing, these two techniques should be jointly enabled to improve performance.

\subsection{Sensitivity to $\alpha$}
We investigated the sensitivity of the hyperparameter $\alpha$, which balances $\mathcal{L}_1$ and $\mathcal{L}_2$, under settings where the extra vision encoder and the padding mechanism are applied individually.
Across configurations, we observe a consistent trend: similar to Table~\ref{tab:loss}, performance improves as $\alpha$ increases from 0.0 to 0.2 and then declines when $\alpha$ becomes larger. This pattern is stable regardless of whether the extra vision encoder and padding are enabled individually or jointly.

Overall, these results suggest that $\alpha=0.2$ provides a robust balance between the contrastive objective ($\mathcal{L}_1$) and the auxiliary supervision ($\mathcal{L}_2$). A moderate auxiliary weight improves the quality of pseudo-visual embeddings and supports cross-modal alignment, whereas an overly large $\alpha$ over-emphasizes auxiliary optimization, which may improve generation fidelity but weakens the discriminative properties needed for retrieval-oriented embedding learning.

\section{Qualitative Analysis of Failure Cases}
\label{app:qualitative}

To better understand the limitations of the modality-completion module, we evaluated UniMoCo in real-world scenarios and analyzed representative failure cases, which can be broadly attributed to two factors: context mismatch and semantic ambiguity.

  \textbf{Context Mismatch} often occurs because textual descriptions do not provide the fine-grained details available in real visual inputs. When the input text is simple (e.g., ``an apple''), the T2I model tends to rely on learned priors and generates a canonical, object-centric view (e.g., a fruit on a white background), whereas real images frequently include substantial environmental context; if the resulting pseudo-embedding lacks this context, it can deviate from the real image embedding that encodes both the object and its surroundings. Moreover, \textbf{Semantic Ambiguity} arises because natural language often contains polysemous words. Although a real image typically resolves such ambiguity through visual appearance, the T2I module must select one interpretation based solely on text, and an incorrect choice can produce a pseudo-embedding that is semantically misaligned with the query.

We illustrate these issues using two image-classification scenarios in which the model failed to retrieve the correct label. In both cases, the query text is identical: ``Please represent the given image for classification.''; specifically, we first present a context-mismatch case and then a semantic-ambiguity case.

\paragraph{Case (a): Context Mismatch}
In this scenario, the query image shows a red apple growing on a tree branch and surrounded by leaves. The correct target text is ``Apple'', while a distractor target text is ``Tree'', and the model incorrectly predicts ``Tree'' instead of ``Apple''. This error occurs because, when processing ``Apple'', the T2I module tends to follow a learned prior in which apples appear as isolated fruits with minimal background, so the generated pseudo-embedding mainly captures generic ``apple'' features; in contrast, the distractor ``Tree'' produces a pseudo-embedding rich in ``leaf'' and ``branch'' cues. Since the real query image is visually dominated by the surrounding context (leaves and branches), its embedding becomes closer to the ``Tree'' pseudo-embedding than to the isolated ``Apple'' pseudo-embedding, leading to misclassification due to background drift.

\paragraph{Case (b): Semantic Ambiguity}
In this classification task, the query image depicts a \textit{baseball bat}, while the target text is simply ``Bat''. The retrieval fails because, when prompted with this polysemous word, the T2I model generates a pseudo-visual embedding corresponding to the \textit{mammal} bat; we infer that this behavior is influenced by the higher frequency of the animal sense in the training data. As a result, the pseudo-embedding is semantically aligned with animals rather than sports equipment, placing it far from the query image embedding in the shared space.

\end{document}